# Leaf Only SAM: A Segment Anything Pipeline for Zero-Shot Automated Leaf Segmentation

Dominic Williams[1*], Fraser Macfarlane[1], Avril Britten[2]

[1]James Hutton Institute, Dundee, UK

[2]James Hutton Limited, Dundee, UK

[*]Dominic.Williams@hutton.ac.uk

## Abstract

Segment Anything Model (SAM) is a new "foundation model" that can be used as a zero-shot object segmentation method with the use of either guide prompts such as bounding boxes, polygons, or points. Alternatively, additional post processing steps can be used to identify objects of interest after segmenting everything in an image. Here we present a method using segment anything together with a series of post processing steps to segment potato leaves, called Leaf Only SAM. The advantage of this proposed method is that it does not require any training data to produce its results so has many applications across the field of plant phenotyping where there is limited high quality annotated data available. We compare the performance of Leaf Only SAM to a Mask R-CNN model which has been fine-tuned on our small novel potato leaf dataset. On the evaluation dataset, Leaf Only SAM finds an average recall of 63.2 and an average precision of 60.3, compared to recall of 78.7 and precision of 74.7 for Mask R-CNN. Leaf Only SAM does not perform better than the fine-tuned Mask R-CNN model on our data, but the SAM based model does not require any extra training or annotation of our new dataset. This shows there is potential to use SAM as a zero-shot classifier with the addition of post processing steps.

## Introduction

One of the main challenges facing plant breeding is that of plant phenotyping [1, 2]. That is the determination of plant performance and characteristics whilst plants are growing. Continued advances in genetic technologies have reduced genotyping costs for plant scientists and breeders, enabling increasingly large datasets to be generated [3]. It is important for advances in plant phenotyping techniques to be made at a similar rate to enable an understanding of plant behaviour and provide data to help understand the physiological impact of genetics. Plant imaging is one of the techniques that can be used to do this and, combined with advances in computer vision techniques, can provide data on plant performance that can show how different genotypes response to stress conditions [4-6]. This paper investigates the problem of measuring potato plants and relating imaging data to leaf area and leaf mass measurements.

There have been ongoing leaf segmentation (LSC) and counting (LCC) challenges over the past several years [7]. Various instance segmentation models have been applied to these images and Mask R-CNN [8] has been shown to perform well in such tasks [9]. Detectron2 [10], offers a framework for applying Mask R-CNN using various backbone region proposal networks and is used in this paper to compare the results of Leaf Only SAM in leaf segmentation tasks to a trained instance segmentation model. There have also been a number of studies trying to expand the generalisability of models produced for the LCC and LSC to other plant crops by using image generation methods [11] or using domain adaptation methods [12]. We have investigated whether Segment Anything can be used to produce a segmentation method in a new crop without the need for training and fine-tuning as an alternative solution to the generalisation problem.

The recently released Segment Anything Model (SAM) presents a "foundation model" that can be used to carry out an annotation free segmentation [13] and it has performed well in a variety of applications. There are several ways it can be used; to generate impressive segmentation results with limited user prompts; to generate highly accurate object masks either from a series of positive or negative points; or to go from bounding boxes to object segmentation [14]. It can however be used as an automatic segmentation technique on its own without any additional user input. A number of studies have been published which utilise SAM for various medical image analysis problems [15-17]. One weakness of many of these methods is that SAM cannot determine the class of the objects being segmented and will segment everything in an image detecting background objects as well as objects of interest. Some early studies have ignored this problem and have instead compared performance of the model by comparing the closest detected object to each ground truth object. This is not possible in real world settings. This limitation can be overcome by applying post processing techniques to the output to evaluate the objects detected and only keep the ones that are of interest. For example one study has used SAM to detect craters on Mars by applying a Hough transform to the detected objects to only keep circular objects [18].

Segment anything used data from many sources online including the leaf counting and segmentation challenge dataset which was used to evaluate the performance of the model [13]. So, this is not an unseen problem for the segment anything model. The fact we have used a novel dataset not previously publicly available for this work ensures that the segment anything model has not had previous sight of the images we have used and highlights the adaptability and generalisation of the proposed approach.

In this paper we present new data with manual annotations on images of potato leaves. This presents a similar challenge to the leaf segmentation challenge, but we have included additional data on leaf area, leaf fresh mass, and leaf dry mass that provides an additional focus to the dataset and ensures that we are evaluating performance that closely links to relevant problem to be solved. We also encounter the limitations of image data collection which itself is not a perfect measure of the biologically relevant traits of leaf area and mass. We present a pipeline that uses Segment Anything with the addition of novel post processing steps as a zero-shot segmentation method. We compare performance of this approach with a Mask R-CNN trained on a subset of our data to see how our method compares against fine-tuned state of the art image segmentation models.

# Methods

## Plants and Imaging

A total of 106 potato plants were grown in two batches. The plants were propagated in tissue culture and then planted into 10x10cm pots and grown in a glasshouse. The first set of plants were 57 potato plants of variety Desiree. These plants were grown in 4 trays of 15 plants with the last tray missing 3 plants which did not survive transplanting into soil. Once a week, each plant was photographed with a DLSR with a top-down shot taken roughly 80cm above the plants which were placed on a common paper background. Each week 12 plants were harvested with three plants being taken from each tray. The harvest plants had their total leaf area measured in cm2, the number of leaves was counted, and the fresh mass of the leaves was weighed. The leaves were then placed in an oven at 50°C for 7 days and then the dry mass was weighed thereafter. Fresh mass can be highly variable with the time since last watering occurred so dry mass is generally favoured as a measure for plant growth. After 5 weeks all of the plants were harvested, and the experiment was complete. A second set of plants consisting of 49 potato plants of variety Voyager was planted three weeks later than Desiree and the same process was applied but with 10 plants being harvested each week instead of 12.

A total of 318 images of potato plants of a varying age between 1 week and 6 weeks growth were gathered. 128 images were manually annotated using the labelme software [19], with a series of points being marked on the leaf boundary to segment each leaf into individual polygons. The annotated images were from the second and third week of imaging and consisted of 45 images of one week old Voyager plants, 34 images of three week old Desiree and 49 images of two week old Desiree. For 32 of these images the plants were then harvested, meaning corresponding ground truth leaf number, leaf area and leaf mass data for these plants is available. To create our segmentation model this dataset was split into random, training (80/128), validation (32/128), and test (16/128) data sets. This resulted in 990 labelled instances of leaves in the training set and 282 and 199 in the validation and test sets respectively. Since no training was carried out for the Segment Anything Model, both the training and validation data sets were used in model development but only the test set is used for evaluation so a comparison can be made with the Mask R-CNN model. Figure 1a shows an example image of the canopy of a potato plant and Figure 1b shows the labelled leaf polygons.

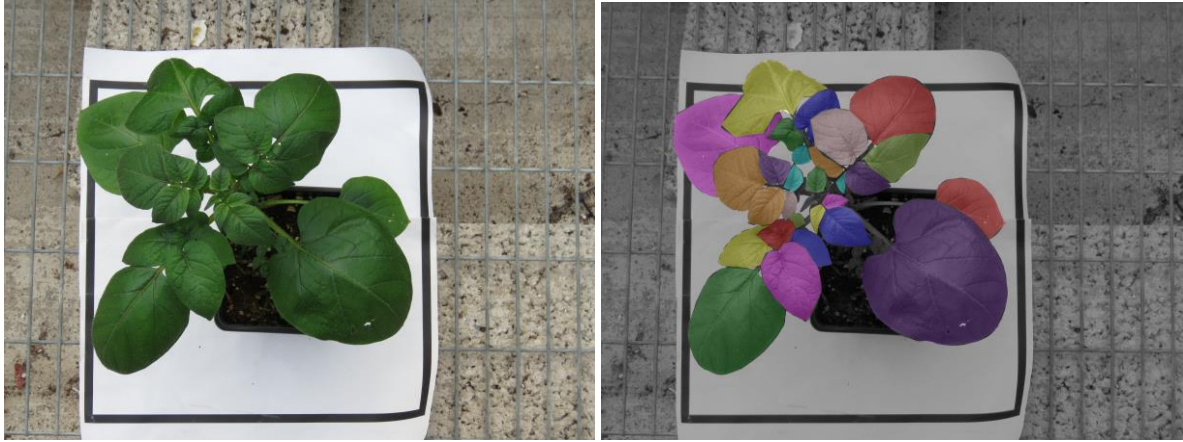

Figure 1: Example a) Image and b) Ground Truth Label pair.

## Leaf Only SAM

We first prompted segment anything with a 32x32 grid of points evenly spaced on the image to generate fully automatic segmentation masks. Segment anything has the option to perform multi-level segmentation where the model is also applied to smaller crops of the image to improve performance in detection of smaller objects. We utilised this to also segment at an additional crop layer. This gives an output of a series of segmentation masks for the images. This includes masks corresponding to objects of interest (the plant leaves) but also many other background objects. We refer to this step as Base SAM when carrying out comparisons. An additional four post processing steps were added to the output to identify only leaf objects.

The first post process step was a colour checking step. This utilises the fact we know we are looking for green plant material so finds green masks in the image. The original RGB images were converted to the HSV colour space. The mean hue and saturation were then used to determine if the objects found were leaves or not by applying thresholds to these values. A mean hue of between 35 and 75 and saturation over 35 were used. We refer to this step as Green pixels when carrying out comparisons.

One of the problems SAM suffers from is ambiguity about the level of object wanted. In our case a segmentation of the entire plant is also a valid object and was often found. A check step was then introduced to remove this if present. If more than two masks were found for an image after the colour check was applied, then a total mask of all pixels was generated. If any objects had an Intersection over Union (IoU) of more than 90% with this mask, then they were assumed to contain the entire plant and so were removed from our list of leaf objects. We refer to this step as Not all when carrying out comparisons.

A small number of other miscellaneous objects were still detected at this point. These were clearly not leaves due to their shape and as a result a shape filter was used to remove these objects. For every mask, the area of the mask was compared to the area of the minimum enclosing circle. If the ratio of mask area was less than 0.1 of the area of minimum enclosing circle the object was clearly not leaf shaped and so removed from our list of leaf objects. Since we wish to detect partially occluded leaves and there is reasonable diversity in leaf shape this step could not be too prescriptive on shape wanted. We refer to this step as Correct shape when carrying out comparisons.

There were still some objects that were present that were a collection of multiple different leaves. We often detected both individual leaves and a mask containing several leaves covering the same area. To remove multi leaf masks we detected multi leaf objects by a simple sum of all the mask objects in the image - labelling each pixel by how many masks it was part of. Any mask with a mean score of over 1.5 was assumed to be a duplicate object. These duplicate masks were then checked to see if they were 90% contained in other masks indicating they were in fact a leaf. Masks that were not contained in other masks were then checked to see if at least 90% of their area was contained in other masks and removed if this was the case.

### Mask R-CNN

Mask R-CNN [8] remains a popular architecture for performing instance segmentation. A Mask R-CNN approach was developed using the Detectron2 framework to compare with the segmentation provided by the proposed Leaf Only SAM technique. Both 50 and 101 layer ResNet feature proposal networks (ResNet-50-FPN and ResNet-101-FPN) were employed as backbones to investigate the effect of CNN depth on the segmentation of our novel potato dataset and trained over 2000 iterations. Training and inference was performed using a single NVIDIA Quadro RTX 8000 with a batch size of 16 images and where images were resampled to have a maximum side length of 1333 pixels. Additional image augmentation techniques, such as rotation, mirroring, and other transformations, which improve training dataset variability were not employed in this work.

### Evaluation Metrics

In order to evaluate the performance of the two methods applied to our leaf segmentation dataset, a number of key metrics were identified. Average Precision (AP) and Average Recall (AR) are used in assessing models applied to the Common Objects in COntext (COCO) dataset and are used here. Specifically two definitions of each are used, the first where Precision and Recall averaged over a number of IoU thresholds $T \in [0.5 : 0.05 : 0.95]$ denoted as AP and AR, as well as that where $T = 0.75$, denoted as AP0.75 and AR0.75.

In addition to Precision and Recall, the DSC was used. As this poses a binary classification problem of leaf vs. non-leaf, the DSC is equivalent to the F1 score and is calculated for each proposed instance as

$$DSC = (2*TP)/(2*TP+FP+FN), \qquad (1)$$

where TP, FP, and FN are the true positive, false positive, and false negative detections respectively.

For the calculation of DSC for the SAM based methods each ground truth mask was compared to the closest detected object since no classification labels are produced. It therefore measures only the accuracy of the segmentation masks themselves not the ability to determine if they are leaves or not. The performance is evaluated after each of the described steps in turn so the effect of each of these can be seen.

### Results and Discussion

Table 1: Comparison of segmentation performance of Segment Anything with Mask R-CNN models trained on the leaf potato dataset

|  | Model | Backbone | $AR_{75}$ | $AP_{75}$ | AR | AP | DSC |
| --- | --- | --- | --- | --- | --- | --- | --- |
|  | Mask R-CNN | ResNet-50-FPN | 83.3 | 81.3 | 74.6 | 72.1 | 0.849 |
| Validation | Mask R-CNN | ResNet-101-FPN | 81.9 | 79.3 | 73.7 | 70.4 | 0.84 |
|  | Base SAM | - | 78.9 | 10.4 | 70.5 | 9.5 | 0.792 |
|  | Leaf Only SAM | - | 78.8 | 71.0 | 70.4 | 63.4 | 0.786 |
| Test | Mask R-CNN | ResNet-50-FPN | 78.7 | 74.7 | 69.8 | 65.4 | 0.871 |
|  | Mask R-CNN | ResNet-101-FPN | 75.9 | 70.2 | 68.4 | 63.2 | 0.867 |
|  | Base SAM | - | 64.7 | 12.6 | 59.6 | 11.7 | 0.729 |
|  | Leaf Only SAM | - | 63.2 | 60.3 | 60.3 | 58.3 | 0.700 |

Looking at Table 1 we can see that our Leaf Only SAM model is not able to perform as well as a fine-tuned mask R-CNN model. We achieved an $AP_{75}$ of 60.3 and $AR_{75}$ of 64.7 which while not poor scores are less than the $AP_{75}$ of 74.7 and $AR_{75}$ of 78.7 achieved by Mask R-CNN. This is not surprising since our model had not been trained on similar potato images like the Mask R-CNN model. We can also see that the post processing steps introduced in Leaf Only SAM are important in improving the precision of the model. Base SAM achieves an AP of only 12.6, the recall of Base SAM is slightly higher than the recall of our model but a 1.5 reduction in recall is a good trade off for a 47.7 increase in precision. The DSC of SAM and our Leaf Only SAM, which measures the accuracy of the best fit mask to each ground truth object, shows a worse performance compared to Mask R-CNN indicating that fine-tuned models can still outperform general foundation models like SAM. It may be possible to improve results of SAM by fine tuning the model more heavily on leaf data.

Table 2: Results of ablation study showing the relative performance of our different post processing steps

|  | $AR_{75}$ | $AP_{75}$ | AR | AP | DSC |
|---|---|---|---|---|---|
| Base SAM | 64.7 | 12.6 | 59.6 | 11.7 | 0.729 |
| Green Pixels | 63.7 | 54.8 | 58.8 | 51.7 | 0.718 |
| Not all | 63.7 | 59.7 | 58.8 | 56.0 | 0.700 |
| Correct Shape | 63.7 | 59.9 | 58.8 | 56.2 | 0.700 |
| Remove multi-leaf Objects | 63.2 | 60.3 | 58.4 | 56.4 | 0.700 |

The results for the different steps in our Segment Anything model, as displayed in Table 2, show the importance of adding additional post processing steps. Each line refers to an additional step being added as described in the methods section. Segment anything alone achieves an average precision of only 12.6. Each of the post processing steps we have developed increases precision. The first step removing objects of the wrong colour has the biggest effect, but further increases are seen with each step. There is a slight reduction in the recall of the method with the additional steps. This is a result of them removing found objects. The first step removing pixels of the wrong colour has biggest reduction most of the other steps had no reduction in recall until the final step which slightly lowers recall indicating instances it is removing correctly segmented leaves.

In order to determine what is causing the remaining low precision an analysis was carried out of the false positives masks generated by our Leaf Only SAM model. The was done by manually looking at the outlines of each of the false negative masks plotted on the original image. These were then put into one of 5 categories leaf (i.e., judged to be a true positive), multiple leaves together, only partial segmentation of leaf, a part of the plant that is not a leaf and finally objects which are not part of the plant. Any leaves that were occluded, and so are only partially visible, were classed as a leaf if all of the visible leaf was segmented. Figure 2 shows examples of the segmentations obtained using a) Base SAM, b) Leaf Only SAM, and c) Mask R-CNN. The yellow outlines in Figure 2b indicate a false positive detection. We can see that these are a combination of multiple leaf objects, objects missed from manual segmentation and some small objects which are inaccurately segmented.

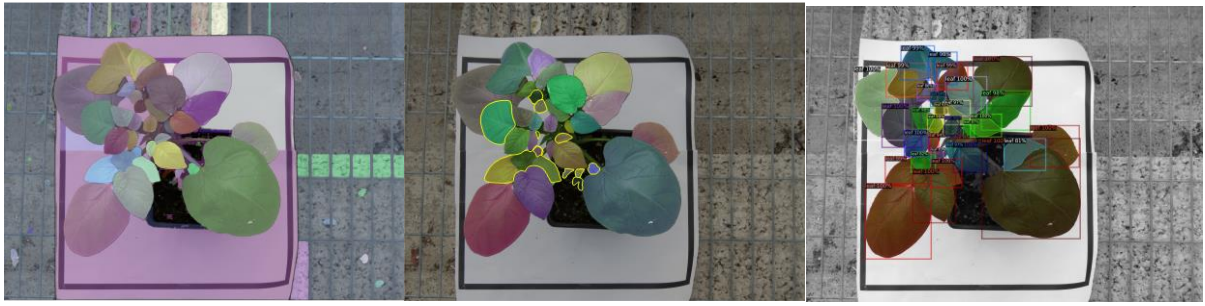

Figure 2: Leaf segmentation on the image from Figure 1 using a) Base SAM. b) Leaf Only SAM. c) Mask R-CNN

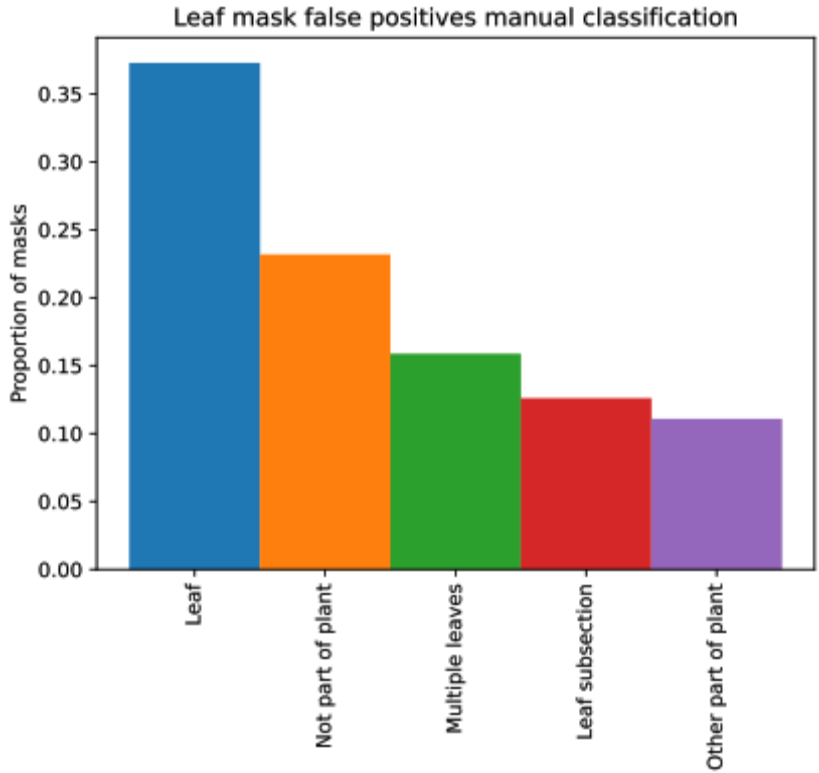

Figure 3: Results of manual classification of false positives from manual classification data.

Figure 3 shows the results of this false positive analysis, we can see 37% of the false positives were judged to be actually leaves. The evaluation was not done side by side with manual annotations so some of these objects will have failed the false positive step due to not reaching the 75% IoU threshold and can therefore be thought of as correct but poor segmentations. Other false positive leaves will be those leaves which are very small or heavily occluded so were missed by manual annotation, the mean size of masks in this category was 18,500 pixels compared to 34,900 for all found masks. This shows the value of using SAM to improve manual data labelling. 23% were of things not part of the plant. The remaining 40% were of plant material but not leaves. There were more masks containing multiple leaves than mask containing only parts of leaves, but both categories were found. A model that was fine tuned on potato plants may be more able to judge where the boundaries of a full leaf are so avoiding these problems.

In order to help understand how our segmentation technique can be related back to real world plant measurements, a comparison looking at both leaf area and leaf dry mass was made. The correlation was calculated between pixel counts on both the manually annotated images and automatically segmented images with the leaf area and leaf dry mass measures. These results, which can be seen in Table 3, show that there is good agreement between manually annotated pixel counts and both leaf area and dry mass r<0.9. The relationship with our automatic segmentation method was weaker r=0.74 and r=0.625 respectively. The relationship of physically measured leaf area to the image derived methods is shown in Figure 4. The stronger relationship between the manually segmented data and leaf area compared to the automatically derived segmentation can be seen. This shows that improving the accuracy of the segmentation method could improve the relationship to manual measures.

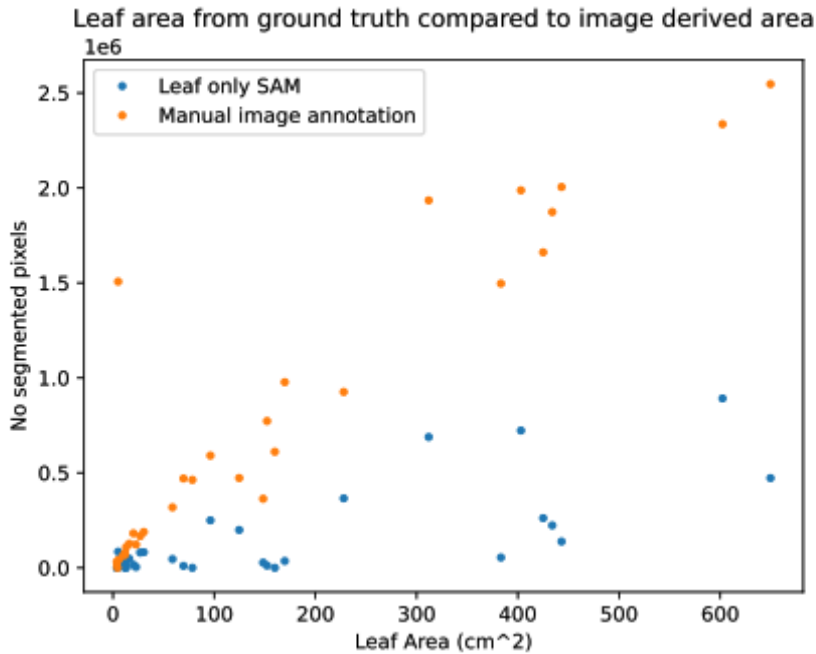

Figure 4: Plot showing the relationship between leaf area physically measured and that from images. For both manual image annotation and automatic derived data.

Table 3: Correlation between physical measures of leaf area and dry mass with image derived measurements.

|  | Leaf area | Leaf dry mass | No. pixels manual | No. pixels Leaf Only SAM |
|---|---|---|---|---|
| Leaf area | 1 | | | |
| Leaf dry mass | 0.891 | 1 | | |
| No. pixels manual | 0.930 | 0.951 | 1 | |
| No. pixels Leaf Only SAM | 0.740 | 0.625 | 0.760 | 1 |

# Conclusions

Our pipeline builds upon segment anything and achieves reasonable accuracy on the leaf segmentation task in potato without any fine tuning or training on potato images. This shows that segment anything is a powerful tool that has potential to be used in the field of plant phenotyping. The removal of the need to have access to annotated data for model training would speed up adoption in more minor crops or growing settings.

There was a reduction of just over 10% for both precision and recall when compared to a fine-tuned model with a slightly larger reduction in dice score. Comparison with leaf area and leaf mass shows that improvements in leaf segmentation techniques could lead to improved relationship with manual data.

Further work could be done in improving the post processing steps. The inclusion of a small CNN based classifier for the masks generated by SAM, similar to the classification branch of Mask R-CNN, could also be another way to improve performance.

## Acknowledgements

This work was supported by strategic research programme (2022-2027) funding from the Rural and Environmental Science and Analytical Services Division of the Scottish Government.

## Data and Code Availability

The data associated with this paper, which consists of images, image annotations and manual measurements, can be found on Zenodo here [20].

The code for Leaf Only SAM can be seen on Github here.